# Minimum Cost Polygon Overlay With Rectangular Shape Stock Panels

**Wilson S. Siringoringo MCIS and Andrew M. Connor PhD CEng**
Software Engineering Research Lab, AUT
Auckland, New Zealand

**Nick Clements and Nick Alexander**
Bisco Ltd
Auckland, New Zealand

Minimum Cost Polygon Overlay (MCPO) is a unique two-dimensional optimization problem that involves the task of covering a polygon shaped area with a series of rectangular shaped panels. This has a number of applications in the construction industry. This work examines the MCPO problem in order to construct a model that captures essential parameters of the problem to be solved automatically using numerical optimization algorithms. Three algorithms have been implemented of the actual optimization task: the greedy search, the Monte Carlo (MC) method, and the Genetic Algorithm (GA). Results are presented to show the relative effectiveness of the algorithms. This is followed by critical analysis of various findings of this research.

## Introduction

In manufacturing industries, the optimization of material plays important part in minimizing the production cost, which in turn contributes to attaining competitive edge for an organization. The importance of material optimization is especially evident in manufacturing goods consisting of large amount of two-dimensional material components such as sheet metal or fabric.

The causal relationship between optimized use of raw material and low production cost applies in the civil construction industry. Various components of a building are covered with rigid sheets cut from stock material, the waste of which is either technically impossible or uneconomical to recycle or reuse. In many cases, the effort involved in cutting the material also contributes significantly to the manufacturing cost. Planning the sheet layout for a section of a building is a tedious process where manual calculation is either impractical or uneconomical, particularly when relatively inexpensive material is used. As a result, builders often allocate material based on loose guidelines only, incurring more cost in acquiring the material as well as consuming more manpower resources for material transport and handling.

A unique variant of the two-dimensional layout optimization problem is found in the residential construction industry. A polygon shaped area such as wall or ceiling is to be tiled with covering sheet material such as Gibraltor board, corrugated sheets or ceramic tiles. With such tiling, it is essential that the entire surface is covered with no gaps or overlaps. The panels are obtained from the supplier in fixed size rectangles. Typically the individual panels are much smaller than the area to be covered. It is also anticipated that the enclosing area may have an irregular outline.

To keep the construction expenses under control, the builder must arrange the panels in a way that keeps the cost variables low. Such parameters include the number of panels allocated, the amount of discarded off cuts, and the amount of effort required for cutting the panels. When the panels are homogenous, such as with sheet metal, it is desirable to reuse the off cuts to cover irregular regions at other places, as this has the potential to reduce the total number of sheets required. A particular example is the reuse of off cuts from corrugated iron roofs (Sibley-Punnet & Bossomaier, 2001). The justification for such effort is provided by the high cost of delivering the roofing material.

The diversity of materials used for constructing a building provides no guarantee that such homogeneity exists for materials used for a particular area. The implication is that the constraints for a particular section of the building cannot be predetermined. In response, a computer program used to resolve such problems must be capable of finding the solution under varying sets of constraints to allow it to be used for any specific instance of the general problem.

Whilst the scope of this paper is of primary interest to practitioners and researchers in the field, the significance of material optimization from an educational perspective should not be ignored. A computer based solution to layout optimization allows students to visualize different solutions and the impact of different design decisions and constraints. Whilst not the focus of this paper, such a visualization tool could be considered an asset in many courses, particularly as the importance of minimizing waste and reducing carbon footprints becomes more significant. A great understanding of the potential for more effective use of materials in graduates would improve the uptake of such technologies in practice.

*Motivation*

Apart from reducing the waste and reducing the associated cost, automating the panel placement design also greatly assists the builder in calculating the required material. When the calculation is done by hand, the common practice is to have a human expert work on the layout and to estimate the number of panels needed to cover a particular part of the building. A few extra panels must then be provided to anticipate the error in the calculation. As the solution only applies to a particular part of the building, the work must be repeated for all other parts as well. The process becomes more tedious when different sizes of the panels are available to choose from. Exploring more than a few different configurations by hand is therefore impractical.

Another inherent problem in MCPO is the lack of guarantee that optimum solutions in the earlier phases will lead to an optimum solution of the entire problem. Coupled with the absence of *a-priori* knowledge about the cost of the subsequent phases, exploring the less-than-optimum early-phase solutions becomes a necessity. Viewed in this light, making the process automatic offers the potential of discovering better solutions than that obtained by hand. When computers are used, more possible solutions can be explored both for individual parts of the building as well as the sum of all those parts. The desired effect is that by providing the raw information to the software, in the form of a CAD model, the builder obtains a detailed and accurate plan about the number of panels required and how they should be cut and placed for the whole structure.

*Related Work*

Considerable research has been done in various fields of two-dimensional layout optimization problems due to the practical needs of industry. Dyckhoff (1990) makes an attempt to provide a systematic classification of such optimization problems. He uses the term cutting and packing (C&P) as a generic name for the problem and all its variants. He further postulates that there are four properties of each problem which determine to which class it belongs. The properties are the problem dimensionality, the kind of assignment, the assortment of the containers and the assortment of the pieces.

Dyckhoff also asserts that there exist 96 classes of C&P problem that result from the combination of the four characteristics. For the purpose of this study however, only the most important variants are considered. The significance of such variants is evident by the amount of research done and the publications that follow. The majority of such problems can be modeled in one of the four main variants: sheet layout, bin or strip packing, rectangular floor planning and the cutting stock problem.

Because most of the research efforts are driven by the need to solve real-life problems, they tend to focus on specific instances of cutting and packing problems. Consequently, the solution approaches are often very closely coupled with the problem being studied, leading to exotic algorithms that are potentially difficult to adopt anywhere else. The most generic and unrestricted form of two-dimensional layout optimization is the sheet layout problem (Lamousin & Waggenspack, 1997), which is also commonly known as sheet nesting or polygon containment. Essentially, the sheet layout problem calls for packing as many polygon-shaped pieces within a polygon-shaped container without any restrictions apart from the basic requirement that the pieces should never overlap. The pieces are allowed to rotate, translate, and to flip about any axis.

Although the generic definition allows the use of arbitrary shaped containers, in practice most problems are characterized by regular-shaped containers such as rectangular sheets (e.g. metal plates) and fixed width with infinite-length source (e.g. fabric or paper). Strip packing (SP) and bin packing (BP) are specific subclasses of the generic sheet layout problem, with the objective limited to placing rectangular items within fixed width container. Furthermore, rotation is allowed only at 90º increments whereas mirroring is irrelevant because of the rectangle's symmetry. The subject of SP and BP covers problems of various dimensions. However, two-dimensional BP and SP problems can be considered a subset of the sheet layout problem class. Lodi et al (2002) define the SP and BP problems respectively:

1. Two-Dimensional Strip Packing (2SP): for a given set of rectangles, a single bin with fixed width and unlimited height (called strip) is provided. The objective is to allocate all the items to the strip by minimizing the height of the strip used.

2. Two-Dimensional Bin Packing (2BP): for a given set of rectangles, an unlimited number of identical rectangular bins of fixed height and width are provided. The objective is to allocate all the items to the minimum number of bins.

The rectangular floor plan (RFP) problem is a finer subset of the sheet layout problem class. With RFP, the problem is limited to arranging rectangle shaped objects within a fixed size, rectangular shaped container. Therefore RFP can be regarded as a special case of 2BP, where the objective is to put as many non-overlapping objects as possible inside a single bin. In the past, RFP has a range of applications such as in metal fabrication and publication layout (Imahori et al, 2005). However RFP later found an application in the design of very large scale integrated circuit (VLSI) chips (Hakimi, 1988; Hsu & Kubitz, 1988; Kiyota & Fujiyoshi, 2000; Murata et al, 1995). Hence despite its being a very small subset of the sheet layout problem, RFP has become an extremely important subject of research in recent years

## Problem Modeling

The polygon overlay problem is found to be composed of two sub-problems which must be resolved sequentially, even though each sub-problem still belongs to the same class of two-dimensional layout optimization. For a given enclosed area and a given dimensions of rectangular panels, the requirement is twofold:

1. Find the optimum arrangement of whole panels in which the covered area within the enclosure is maximized. The by-product of this process is a set of irregular shapes which represent the remaining exposed areas.

2. Resolve how such irregular shapes can be nested within the minimum number of panels. Shapes that are bigger than the panel itself are cut at angles parallel with the rectangle's axes to allow such nesting.

This decomposition into two sub-problems can potentially mask the complexity of the task of finding the optimum solution. It is important to recognize that in the construction industry, the actual size of the panels is in itself a design parameter. In some applications, the panel size will remain fixed for the two sub-problems whilst for other applications the panel size could potentially be varied. With this in mind, it becomes apparent that the problem is complex with potentially many sub-optimal solutions.

At the end of the calculation process, the desired output consists of numerical and graphical information:

1. The total number of panels, consisting of panels to be fitted whole and the remainder to be cut to produce the irregular shapes.

2. The nesting plan with which irregular shapes are cut from whole panels.

3. The area overlay plan with which whole panels and irregular cuts are fitted to the enclosed area.

It is important to note that although the two sub-problems are similar, they are resolved with mutually unrelated and potentially conflicting objectives. As an example, the lowest cost for first sub-problem may be to cover as much area as possible with the least number of panels. However, the optimum solution second sub-problem may be the least amount of cutting. Hence a cheap solution in the first phase may lead to expensive penalties in the second.

## Optimization Algorithms

### Placement Strategies

Because of the limitation of computing resources, early solutions to sheet nesting problems were based on sequential placement of the pieces. The availability of more powerful computers has made possible the approach of simultaneous placement of the nested pieces. The simultaneous placement approach allows for wider exploration within the search space, which increases the chances of finding better solutions than that obtained from sequential placement.

Sequential placement algorithms are characterized by populating the container with one piece after another. When a piece is placed on the container, an irregularly shaped smaller container is in effect created. The algorithms greedily conserve the size of the newly created restricted area as it picks subsequent pieces. The process is repeated until either the pieces are exhausted or the container is unable to accommodate more pieces. An equally important aspect of sequential placement algorithms is the optimization of coordinates and orientation of the pieces. Linear programming is perhaps the most popular approach found in the literatures. With linear programming, possible coordinates and orientations are limited to discrete values only. The configuration that yields the optimum value for the subsequent objective function is then selected. Laurent and Iyengar (1982) provide an example of linear programming in use for solving nesting problem with rectangular objects.

With simultaneous placement, pieces are paired and placed in the container without using any sequence allocation list. Instead, other data structures such as trees and graphs are used to represent the nesting and the position of each piece relative to one another (Bounsaythip & Maouche, 1997; Bounsaythip et al, 1995). The optimization task is accomplished by finding the configuration of such structures which provides the best value for the objective function. Typically the there is a very large number of possible configurations for a given nesting problem, which makes an exhaustive search unfeasible. Several researchers argue that the sheet nesting problem generally falls into an NP-hard computational complexity category. Faina (1999) concludes that the implication of being an NP-hard problem is finding the absolute optimum is not feasible when the number of items is large.

### Greedy Algorithm

Solving optimization problems typically involves the process of going through a series of steps, making decisions from a set of possible choices at each step. If the information about payoff for each choice is available, such optimization problems can be solved using relatively unsophisticated methods such as a greedy algorithm. At any point, the greedy algorithm always picks a choice that gives the best reward at the moment. No consideration is given for the lesser immediate payoff alternatives, despite the potential of greater long-term reward. Because of this characteristic, greedy algorithms are simple in concept and easy to implement.

There is a major weakness of this strategy which is the inability to escape local optima. In a classic hill climbing problem, the algorithm makes an ascent by successively selecting the highest neighboring node until the peak is reached and no more climbing is possible. Obviously, this approach is prone to premature convergence if the search space happens to contain multiple local optima.

Greedy algorithms seldom find the global optimum solutions, yet in many cases they are capable of finding reasonable solutions quickly (Cormen et al., 2003). Because of its simplicity and speed of execution, the greedy method is quite powerful and well suited for a range of problems. Greedy methods are used in a number of important algorithms such as minimum-spanning-tree algorithms, Dijkstra's single-source-shortest-path, and for data compression using Huffman codes (Cormen et al., 2003).

In the optimization domain, greedy algorithms may not be the best choice of algorithm because they cannot reliably find better-than-average results. Nevertheless, a greedy algorithm implementation is important for this research for a number of reasons. Firstly, it provides an easy to construct platform to verify the correctness of the problem modeling. More importantly, however, it serves as the baseline solution against which the performances of more sophisticated algorithms are measured. For industrial use, it is important to trade off solution quality and speed of convergence, and it may be that for the building services industry that the greedy algorithm baseline may provide sufficiently good solutions in an acceptable timeframe.

## *Monte Carlo Methods*

A heuristic approach is commonly used in optimization problems when the search space is too large for exhaustive exploration. In a heuristic algorithm, rules and methods are applied to narrow the search towards most promising areas of the search space (Dean et al., 1995). Heuristic techniques are a major subject in the field of Artificial Intelligence (AI) as many such problems can be represented as a large search space, from which the solution is to be discovered. Heuristic techniques exist in many forms, and are the key ingredient for many successful and robust AI algorithms. In the Genetic Algorithm discussed below, for example, the guidance of the search towards promising areas takes the form of the three genetic operators of selection, crossover and recombination.

Despite being effective in general, heuristic techniques do not guarantee success in every case. The do however, offer better chances of good result most of the time. In some cases, particularly when the objective function has a discontinuous or random pattern, a blind guess may give equal or better results than the guided search (Dean et al., 1995).

Monte Carlo (MC) method is a blanket term used to describe any method characterized by the use of a random number generator and the complete disregard of dynamics involved in reaching the results. Weisstein (1999) defines Monte Carlo technique in general as:

> "… any method which solves a problem by generating suitable random numbers and observing that fraction of the numbers obeying some property or properties. The method is useful for obtaining numerical solutions to problems which are too complicated to solve analytically."

Apart from the transition probability, which is constant, decisions at any stage are made without any restriction in a Monte Carlo method. The original Monte Carlo method was first used to create models in statistics, but has since been applied in solving various optimization problems. In its most basic form, a memory-less random walk is all that is involved in implementing Monte Carlo optimization algorithms. Such unrestricted search, completely lacking in decision making rules and record keeping, makes Monte Carlo algorithms much simpler to implement than many heuristic algorithms. The actual Monte Carlo algorithm implementation will be discussed later, but as an example of the simplicity of the algorithm, the implementation only requires two candidate solutions to be maintained (as opposed to a large population) and the manipulation of the candidate solutions doesn't involve anything more than simple manipulations of bits in a binary string.

*Genetic Algorithms*

Evolutionary computation and optimization were born when researchers came up with the idea of developing powerful optimization algorithms based on simulation of evolutionary process. The efforts spawned a number of algorithms, of which Bäck & Schwefel (1996) have identified three mainstream methods: the genetic algorithm (GA), evolutionary programming (EP), and evolution strategies (ES).

These algorithms use the concept of a population of individuals which is subject to a series of probabilistic operators such as mutation, selection and recombination. Each individual represents a potential solution to a given optimization problem. During the computation process, the population will undergo a draconian process in which stronger individuals will thrive while the weaker ones perish. Genetic algorithms, which were first developed by John Holland and his colleagues at the University of Michigan (Holland, 1975), exhibit all the three main characteristics of evolutionary computation (Bäck & Schwefel, 1996). Their research goals were to rigorously explain the adaptive processes of natural systems and to design an algorithm that faithfully replicates the important mechanisms of natural systems.

In a GA, an individual is represented as a string of genes, or chromosome. Unlike its natural counterpart, however, the genes do not manifest themselves in the physical traits of the organism. The algorithm is only interested in the gene string itself as the potential solution of the optimization problem. No mapping to physical characteristics is necessary or desired beyond that which is required to evaluate the fitness of the candidate solution. From the optimization point of view, the chromosome serves as the representation of the coded parameters of the optimization problem. To determine how 'good' an individual is as a solution, its chromosome is decoded to retrieve the actual values, which is then fed to the objective function of the original optimization problem. The routine that decodes the gene string and calculates its objective function is called the fitness function, and the result of the examination is called the fitness value. Gene strings with better fitness values represent the stronger individuals within the population. Such individuals are favored by the system and more likely to survive and reproduce. Genetic algorithms start with an initial population, which will be successively replaced by newer generations until the algorithm terminates either when sufficiently good individual is found or the number of generations has exceed the limit set by the user. Many variants of GAs exist, but they are generally easy to recognize as they are constructed using the same outline.

In many GA implementations in the literature the chromosome is commonly implemented as a finite-length binary vector. A binary vector provides the maximum flexibility for parameter coding and interpretation much the same way as basic data types such as numerical or symbolic values are internally represented in the computer memory. Non-binary strings are also used however, in specific cases such as when representing nodes in Traveling Salesman Problem (TSP), where a binary equivalent is impractical or inefficient (Ansari & Hou, 1997).

Because of its flexibility, coding the optimization parameters into a gene string can be a daunting task. For any given optimization problem, there are typically a number of possible ways to code the parameters into the gene string, and some may be better than others. There is surprisingly little available literature providing guidance for coding GA parameters. Coding guidelines for specific domain do exist however, such as those proposed by Nagao for optimization of numerical parameters (Nagao, 1996).

## Algorithm Implementations

In this work, three algorithms have been implemented to allow a comparison of their respective performance to be made, namely a simple greedy search, a Monte-Carlo (MC) algorithm and a Genetic Algorithm (GA). This section briefly outlines the details of each of the algorithms.

### Greedy Search

For the MCPO problem, the greedy algorithm constructs a solution sequentially by always trying to fit the most profitable piece into the available free space. This is a short-sighted strategy whose performance can be extremely poor in complex solution spaces due to its inability to escape from local optima. Nonetheless, the mechanism of a greedy algorithm is intuitive and therefore easy to implement into reliable code and also provides a baseline performance to be measured against other methods.

While the concept is simple, the implementation in nesting problems is much more involved because the basic greedy algorithm works only with scalar values. To solve a nesting problem, the algorithm needs to be modified to take vector values into account. Vector values differ fundamentally from scalar values in that simple arithmetic operations do not apply. Accommodating vector values in a greedy algorithm proves a non-trivial task.

For any given iteration in a nesting search, the greedy algorithm must resolve four key problems:

1. Which candidate piece to select
2. Where to put that particular piece in the nesting container
3. What orientation the piece should be placed in
4. Whether flipping should be applied for the piece if orientation constraints allow

Most of the corresponding parameters can be represented as scalar values; however the actual location of a piece in the container requires a vector parameter for representation.

The basic greedy algorithm essentially addresses only the first problem, whereas the remaining three are not covered because they are specific only to the domain. Valid answers to those additional three will in effect justify the decision made for the first problem by proving that the piece in question can be successfully nested. Recall that the search for valid answers must not violate the two fundamental constraints of the nesting problem, in that the pieces must lie entirely within the boundaries of the container and must not overlap with each other.

At this point it is also important to appreciate that the objective of the layout optimization is to minimize wasted material. Modeling the problem for a greedy algorithm therefore requires the understanding of how the values of the pieces are quantified to allow the resulting waste to be directly calculated. Given that waste is independent of the thickness of each panel, surface area has been selected as the main parameter for the greedy algorithm search. When a candidate piece is to be evaluated, the surface area of the vacant space in the container is calculated. The algorithm then attempts to fit the biggest piece in the pool whose area is smaller or equal to the vacant space into the container. If this attempt is unsuccessful, the next biggest piece is evaluated. The process is repeated until a piece can be legally fitted into the container, which also results in that piece being removed from the pool. If none of the candidate pieces in the pool can be selected, a fresh container is used and the process is repeated.

The algorithm terminates when all the pieces have been used. Because the pieces are the product of the original layout area when it was cut up according to the shape of the container, there will be at least one way to fit a piece into an empty container. Therefore the algorithm is always guaranteed to terminate. The greedy notion of this algorithm is realized by sorting the pieces based on their surface area in descending order before the actual optimization takes place.

The second problem to be solved about a particular piece is about where it should be placed within the container. The best solution is defined as the one with highest total length of the conjoined edges, as the length of such shared edges is a reasonable approximation of the amount of obstruction in the vacant space. A full examination of placements strategies has been previously published (Siringoringo, 2007).

*Representation for Monte Carlo and Genetic Algorithm*

The Monte Carlo and Genetic Algorithm approaches have a shared problem representation, based on a candidate solution that encompasses a number of pieces clustered into a number of panels, represented as a bit string or chromosome. The two algorithms then manipulate the bit string using different mechanisms to attempt to discover an optimum solution. Details of this representation have been published previously (Connor & Siringoringo, 2007), but are briefly described here.

Substantial effort has been expended in designing the required chromosome for the representation. Not only because there are multiple parameters involved in layout optimization problems, but some of the parameters are also inter-dependent. To construct a suitable model, it is quite worthwhile to examine the parameters that define a second-stage solution in MCPO. Such parameters are:

1. The total number of stock panels required
2. The list of pieces that are nested within each stock panel
3. The placement coordinates of each piece within a stock panel
4. The rotation and flipping applied to that particular piece

Evidently the first parameter is dependent on the second parameter. Similarly the second parameter is largely dependent on the third and fourth parameters. In the face of this, the only information available to determine the value of those parameters is the list of irregular panels represented by their vertices. This all leads to a situation radically different from standard sheet layout problems found in the literature.

To reiterate, in standard sheet layout problems commonly found in the literature, only a single container is provided. The solution designer is therefore allowed to use the chromosome to directly represent the container and map the genes within the chromosome to the nested pieces. Static blocks of bits can be used to represent the placement coordinates of each piece, its rotation, and so on. This static mapping cannot be easily applied to MCPO, since the number of containers itself is a variable to begin with.

A much more feasible solution is to deliberately use only a few parameters in the main model, and to relegate the task of populating the rest of the parameters somewhere else. Since the first two parameters identified above are the most crucial, they are selected to be represented in the chromosome. The solutions provided by both the GA and MC therefore only contain the information about how many stock panels are used and the list of pieces that are nested within each stock panel. The problem of how those pieces are actually nested remains unsolved at this level.

Resolving the third and fourth parameters is important to determine whether the solution for first and second parameters is legal. It is most appropriate to make finding their correct values an integral part of the fitness evaluation function for the original chromosome.

There are two logical ways to solve the above secondary problem at a technical level. The first is by utilizing the same sequential placement routines as used in the greedy algorithm. The second is by mapping the now-static parameters as chromosomes to be processed by the same GA or MC engines used to solve the first two parameters. Direct coding to the genes in the chromosome is still not possible because the second parameter is of a variable length. To solve this problem, indirect coding employing the concept of clusters is used.

In this technique, static blocks in the chromosome are mapped to the pieces to be nested. This represents the worst case solution, where each piece requires an individual stock panel to be used. From the first step of the solution, it is known that all pieces to be nested are smaller than the stock panels therefore this provides an upper threshold for the maximum number of panels required. Each piece to be nested is associated with a fixed-width block of bits in the chromosome. This block contains only a single variable of integer type, namely the cluster ID. Figure 1 shows the association between the pieces and the blocks in the chromosome.

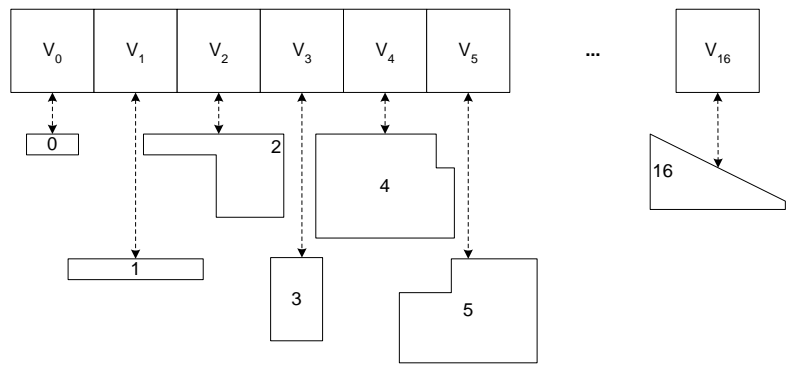

*Figure 1*: Gene to Panel Mapping

The value of each variable points to an imaginary cluster of pieces that will be fitted with in a stock panel. Figure 2 shows an example of a populated chromosome with the imaginary clusters that result. Because only 17 panels exist, the binary string can use five bits to hold the cluster ID.

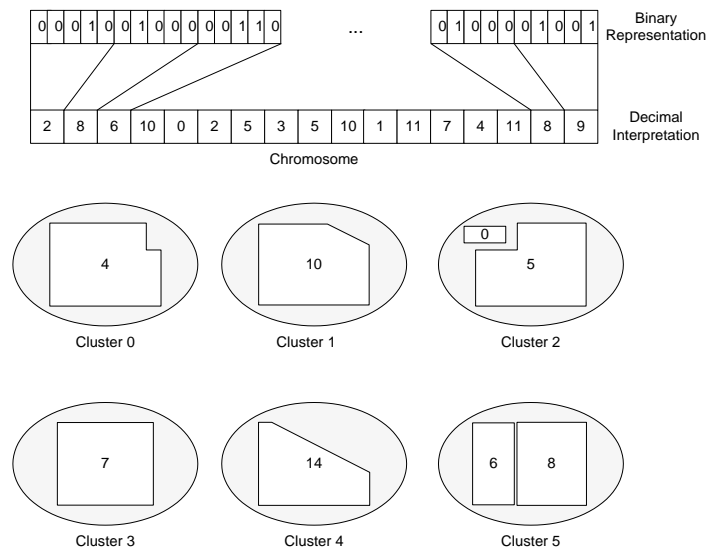

*Figure 2*: Interpreting a Candidate Chromosome

Using Figure 2 as reference, it is easy to decode the chromosome to find that the Piece 0 is a member of Cluster 2, whereas Piece 1 is a member of Cluster 8, and so on. Similarly, Cluster 0 appears to have only a single member, i.e. Piece 4, whereas Cluster 2 has two members: Piece 0 and Piece 5. In the example given, there are a total number of 11 clusters, which if they are all legal (i.e. all the pieces fit with in a stock panel), this gives the total number of stock panels required. This representation does raise issues concerning how invalid clusters are dealt with, and this is discussed in more detail in previously published work (Connor & Siringoringo, 2007; Siringoringo, 2007).

*Monte Carlo Technique*

Simulation with the Monte Carlo technique uses only a pair of bit strings: the working chromosome and the current best chromosome. For the number of iterations specified by the user, the working chromosome is subjected to random manipulation and its fitness value is calculated. Whenever the fitness value of the working chromosome is better than previously found, the bit string is copied to the current best chromosome and the process starts again. The random manipulation for MC is simply done by flipping random bits in the chromosome. The user supplies the numerical constants that control the number of bits that may be flipped, and the probability of a selected bit to actually be flipped.

A fitness value is calculated by considering the total amount of vacant surface area found in the nesting containers. Because the objective of the optimization is to minimize this area, a lower fitness value is taken as the better fitness value. For each nesting plan, vacant surface area is simply calculated as the area of the container subtracted with the total area of all the pieces nested inside.

The MC technique represents an undirected search. The algorithm employs no particular strategy other than exploring the multidimensional search space rather aimlessly by randomly changing direction along certain axes at each cycle in the hope of coming across a good solution. Although there is a certain degree of inertia provided by the unchanged bits, they do not in any way contribute to directing the algorithm towards likely better solutions.

*Genetic Algorithm*

As opposed to the MC technique, the genetic algorithm performs the search in the directions that promise the best result. Instead of just a single working chromosome, a population of chromosomes is used. The search direction is controlled by various bit patterns contained within the population. The GA shares the same chromosome structure and fitness function as those used by MC algorithm.

At the conceptual level, the GA implementation for nesting optimization follows the outline given in Chapter 3. The actual code is based on the simple GA implementation in Pascal written by Goldberg (1989). The basic genetic operators of selection, crossover and mutation are used.

In terms of selection, the most significant attempt at obtaining better future solutions from an existing set of chromosomes is population sorting. When a population is generated, its members are sorted according to their fitness values. Chromosomes with better fitness values are placed higher in the list, implying higher chances of being selected to mate. The chromosomes are then selected in pairs for mating, during which crossover occurs.

A chromosome that has been selected is not eliminated for the selection of the next pair, and stands the same chance of being selected again as before. The reason behind this policy is to allow a supposedly good individual to contribute more than once in creating the next better generation. Population sorting is the key aspect that differentiates this particular implementation of a GA from a completely random search such as the Monte Carlo technique. Without population sorting and the survival for the fittest rule it implies, the GA will degenerate into a series of indiscriminate matings between random chromosomes with no real chance of optimizing the result.

Given the representation used where clusters are mapped to a chromosome, a chromosome may contain a number of good clusters, i.e. clusters that translates into a nesting plan with small waste area, as well as bad or invalid clusters. Leaving the good clusters untouched while actively working on the rest is proposed as a good strategy. Because the existence of the clusters is only implied by the pieces that "belong" to them, the bit pattern of good clusters is immediately found in the bit pattern of the variables within the chromosome referring to them. In other words, the bit pattern of the good clusters is static. Ergo, a more advanced concept in GA associated with the bit patterns, the schemata can be brought into play.

How the schemata can be used to further enhance the GA implementation for layout optimization has not been explored in this project, mainly because of the perceived complication associated with capturing and handling the bit patterns. Such an investigation remains an interesting subject however, and given the potential to improve the performance of the algorithm, further research into the area in the future may prove worthwhile.

## Experimental Results

The number of parameters used in solving layout optimization problems is normally such that the possible number of combinations that can exist is very large. This rules out exhaustive investigation because such a search is computationally very expensive. The experiments therefore need to be configured and executed in such a way that allows the behavior of the software to be monitored and measured through only a small number of optimization runs. With such limitations in mind, the experiments have been conducted with the aim of observing the optimization processes. To reiterate, the objective of layout optimization is twofold:

1. Generate a layout of a set of stock rectangular panels which covers the container region

2. Generate a set of layouts where irregular remaining shapes of the original container can be fitted back into the stock rectangular panels

Two crucial tasks need to be performed by the optimization engine for the first phase: determining the point of origin on which the bottom-left corner of leftmost panel will be placed, and selecting the particular stock panel that returns the most favorable result. The second optimization phase is somewhat easier for the optimization engine because it is only required to search for a set of layout plans according to user-specified parameters.

Trial runs immediately reveal that for a given point of origin used in the first phase optimization, there is a sizeable amount of computation that follows before its corresponding final result can be obtained. This problem is particularly severe when simulation-based optimization algorithm is used in the second phase. As will be discussed in later sections, resolving a moderate-sized problem using simulation-based algorithm for a single point of origin can easily take hours or days even when reasonably powerful computer hardware is used.

A major contributor to the computation cost however, is the multiple candidate stock panels associated with each optimization case. Because selecting the most productive stock panel dimensions from a pool of candidates is one of the prime objective of MCPO optimization, this feature cannot be dispensed with and the resulting computational cost must be accepted.

It is clear that exploring multiple points of origin is not a feasible option except in very simple cases. Real-life examples are typically complicated enough to render multiple points of origin prohibitively expensive to compute, regardless of the strategy in selecting those points. Because of this reason, all the optimization runs will be conducted with a single predetermined point of origin only. The chosen point of origin is at (0, 0) in the workspace coordinates, which is arbitrary because of the non-unique way the container can be placed in the workspace. Future research can explore ways of making intelligent selection of the point of origin.

There is still an array of parameters whose values need to be determined before the second stage optimization can take place. In commercial setting, the users can tune all of these parameters through the UI according to their own preferences and reasoning. In these experiments however, the main interest lies in finding the comparative performance between algorithms in terms of execution time and the quality of the results. Consequently, few of the parameters will change during the course of the experiments. Values set for those parameters and the justification behind them will be provided on per case basis. Apart from the algorithm performance, the experiments will also provide some additional information that may be of importance. Especially interesting is the effect of placement strategy (first-fit compared to best-fit), piece flipping and rotation to the overall efficiency of the solution.

*Simple Rectangular Layout*

The first experiment involves the layout optimization of a 300x300 square container, with a 50x100 rectangle-shaped obstacle within as shown in Figure 3.

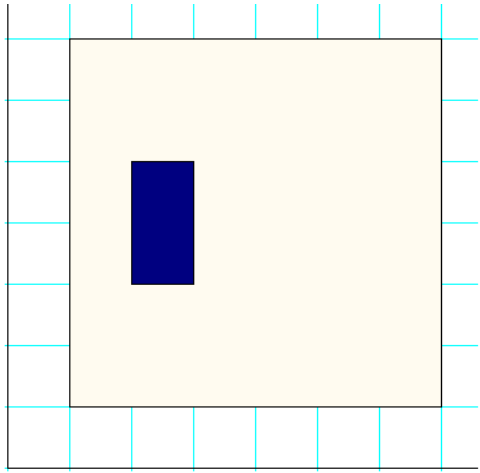

*Figure 3*: Simple Rectangular Container Optimization Problem

The bottom left vertex of the shape is (50, 50). The optimization procedure seeks a solution with which 50x100 rectangular shaped stock panels can be used to cover the container area, using (0, 0) as the point of origin. The origin is outside of the area to be covered and this simple example allows the impact of this to be observed.

This is a trivial example, the purpose of which is to demonstrate that the optimization process is actually able to find a solution for a simple problem. Figure 4 shows the solution of the first part of the problem, whereas Figure 5 shows the solution of the second stage optimization using greedy algorithm. The lighter shade is used to indicate regular, whole panels, whereas the darker shade indicates irregular panels.

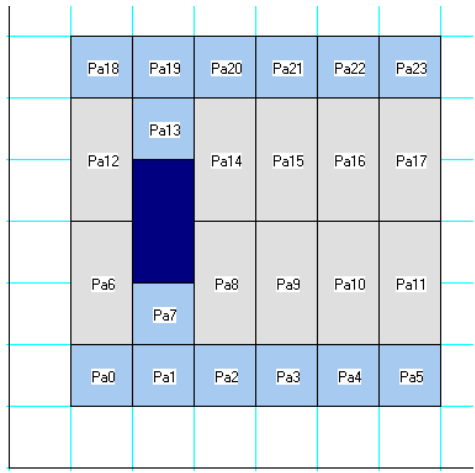

*Figure 4*: Panel Placement Solution for Simple Rectangular Container Problem

In Figure 5, the whole and part panels used in the solution are rearranged to show the total number of whole panels required. In this simple case, each part panel nests perfectly into a number of whole panels.

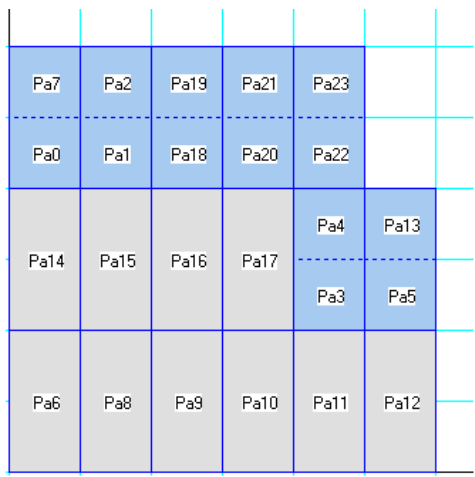

*Figure 5*: Nesting Solution for Simple Rectangular Container Problem

The solution efficiency is defined as the container area divided by the available area provided by the stock panels. All three optimization algorithms prove consistently successful in finding 100% efficiency solution. The simple rectangular container however, is not typical. Solutions with less than 100% efficiency are the norm as subsequent experiments will show.

Despite the 100% material efficiency, it is clear that the solution for the simple rectangular container problem above is not ideal when point of origin (0, 0) is used. The absolute best efficiency is achieved when (50, 50) is used as point of origin instead, as evident in Figure 6.

Only entire panels are used in this case, implying not only 100% efficiency but also the complete absence of cutting the material.

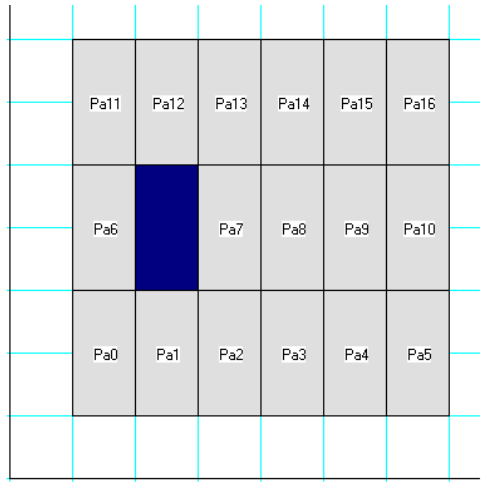

*Figure 6*: Best Nesting Solution for Simple Rectangular Container Problem

It has been explained in at the beginning of this section however, that only a single point of origin (0, 0) is to be used throughout the experiments. Therefore to be consistent with the experiment strategy laid out early in this section, result such as that in Figure 4 is not to be considered any better than that in Figure 1. Future work will focus on the development of the geometric functions required to select a more appropriate origin based on querying the shape of the polygon to be filled to determine the best starting point. This would eliminate the need to conduct multiple optimizations to determine the best origin.

*Single Wall Layout*

The second experiment involves the layout optimization of a single container with both convex and concave corners. As shown in Figure 7, the outline of the container takes the form of the wall at the side of a building. The shape has the height and width of 450 and 350 units of measure, respectively. Assuming that the panels do not have grains or patterns, rotation at 90 degrees increments is allowed during nesting process.

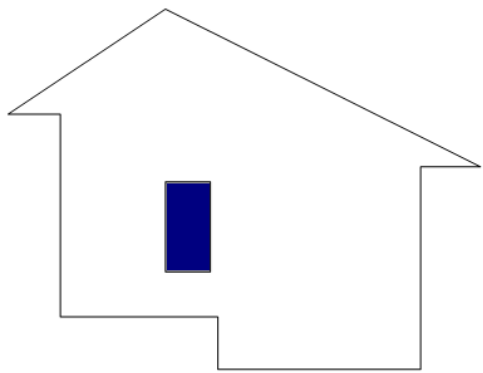

*Figure 7*: Single Wall Optimization Problem

Two types of stock panels are being considered to generate the solution: one has the dimensions of 80x60, the other 54x80. Because the shapes can be easily scaled to their life-size equivalent, there is no need to map unit of measure used in this example to the standards actually used in the building industry. The first stage solution is shown in Figure 8. As with the previous case, it is clear that the use of the (0, 0) origin is affecting the quality of the resulting solution adversely. Being able to identify, without human intervention, a better origin would in this case lead to improving the quality of solution by reducing the cutting required to produce the layout, even if the number of panels were not reduced.

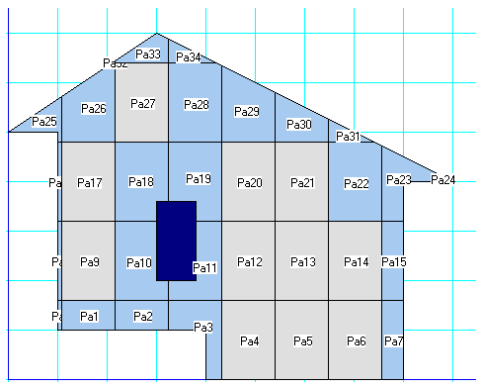

*Figure 8*: Panel Placement Solution for Single Wall Problem

Such improvement would be most evident in the number of pieces required to be nested in the second stage. An example of a second stage solution is shown in Figure 9. In this figure, the whole panels used are shown in the lighter shade, and the component parts, or off cuts, are shown nested into the number of whole panels required. In essence, this figure shows the total number of panels required to cover the wall but without reference to the shape of the wall. Unlike the simple rectangle container problem however, it is not possible to achieve a solution with 100% efficiency.

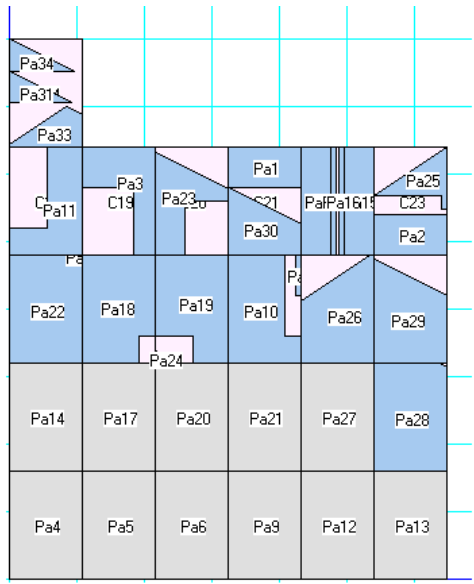

*Figure 9*: Nesting Solution for Single Wall Problem

The greedy algorithm achieved efficiency is surprisingly high at 87.31% in most cases when first-fit strategy is used and 180 degrees rotation is allowed. Because greedy search is deterministic, the result is always identical for a given set of parameters. Therefore, only single optimization run is executed for each parameter set. Optimization using the Monte Carlo method is less successful with the material efficiency of 84.94%. The number of iterations and the flip probability are set to 10,000 and 0.2 respectively. Such numbers have been selected after a series of trial runs to reflect a perceived good combination of moderate-size search with relatively low rate of bit mutation.

The Genetic Algorithm achieves similar performance, achieving 85.42% efficiency in for all parameter settings. In all cases, a population of 100 individuals is used to evolve in 100 generations. Crossover probability of 0.6 and mutation probability of 0.1 are used, also after such numbers appear to be adequate in a series of trial runs. In the absence of better methods for setting the parameters, this approach seems sufficient for our purpose.

Perhaps the most important finding in this experiment is that greedy search outperforms the two other algorithms despite its obvious weakness of being incapable of recovering from premature convergence. The MC and GA on the other hand seem unable to capitalize their advantage in negotiating local optima. The net result is not only the greedy search being capable of finishing the job much faster (one nesting attempt instead of 10,000), but also with better quality result. However, it is important to point out that the total number of panels required for the single wall is the same even though the utilization of material is lower. The savings in material become significant as the application of the approach is extended from the optimization of the single wall, to the room, and ultimately to the whole building. Potential exists to significantly reduce the total amount of material required if the approach can be applied to the optimization of whole buildings. Another important finding is the impact of rotating the pieces to the efficiency of the final nesting result. More freedom of altering the orientation of the pieces does not automatically translate to more efficient nesting solution. Finally, the best-fit placement strategy does not guarantee better solution than first-fit strategy, both in terms of area utilization and shared edge length. This finding is rather unexpected, because the best-fit strategy has been aimed at maximizing the length of the shared edge.

*Simple Roof*

The third optimization problem is a sample problem related to roof layout optimization (Sibley-Punnet & Bossomaier, 2001). In this particular case, multiple containers are used. The simple roof layout differs from the previous problems by the multiple containers involved. Figure 10 shows the top view of the roof. Sections of the roof have been labeled 1-4 to assist identification.

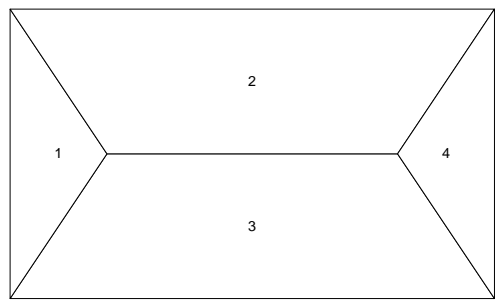

*Figure 10*: Simple Roof Viewed from Above

The simple roof layout also differs from previous problems from the material constraints of the sheets used. Sibley-Punnett & Bossomaier (2001) assume that the panel used takes the form of corrugated iron or similar material. The implication is that the panel has distinct upper and lower sides, rendering flipping illegal. The material also has ridgelines and guttering that dictates that only 180 degrees rotation is allowed.

Yet another constraint to be taken into account in this optimization problem is the overlap between adjacent pieces when installed on the actual roof. Such overlap exist in the actual roof construction both for aesthetic reasons and to prevent leakage. In this example however, such overlap is ignored to avoid unnecessary complication. Very high efficiency of 92.29% is achieved by greedy search. In stark contrast, Monte Carlo method is only able to produce solutions with efficiency of 77.05%. All the parameters have been set identical to that in single wall layout problem previously discussed. The Genetic Algorithm yields disappointing results, achieving efficiency of only 76.40% in its solutions. All GA parameters have also been set identical to that in single wall layout problem.

The superiority of greedy search becomes much more apparent in this experiment. Neither the MC nor GA is able to create solution with efficiency that matches even the lowest of that generated by the greedy search. Figure 11 illustrates the solution with the whole panels used shown in the lighter shade, and the component parts, or off cuts, are shown nested into the number of whole panels required. In essence, this figure shows the total number of panels required to cover the roof but without reference to the shape of the roof.

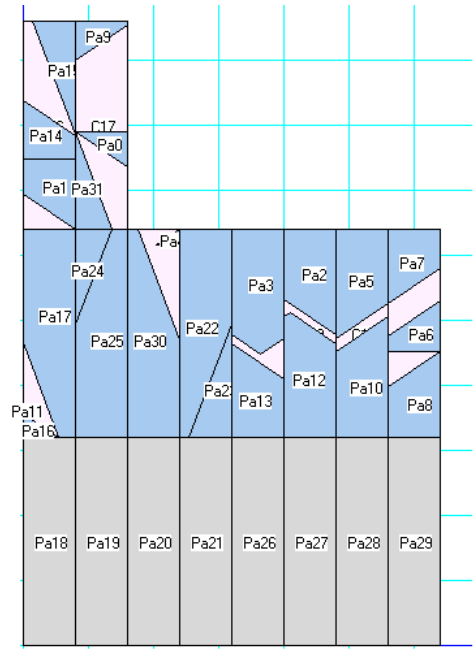

*Figure 11*: Nesting Solution for Simple Roof Problem

As in the previous example, the total number of panels required is the same in each case, apparently not offering the savings that should be possible given the extension to multi-surface optimization. In the majority of the cases, better efficiency is obtained when rotating the pieces by 180 degrees is an option and even better results possible if free rotation is allowed. The constraints on the problem due to the material are limiting the ability of reducing the number of panels required. As in previous experiment, best-fit placement strategy does not provide direct help in achieving better overall efficiency. It does consistently yield better result in terms of shared edge length, however.

## Complex Roof

The fourth and final experiment uses another example from reference (Sibley-Punnet & Bossomaier, 2001). In this case, a complex roof consisting of multiple sections is used. Unlike the simple roof example, there are twice as many sections of greatly varying sizes that make up the roof. Concave shaped sections are also used, as opposed to all-convex shapes in the simple roof layout problem. Other roofing material-specific constraints still apply however. Figure 12 shows the top view of the complex roof.

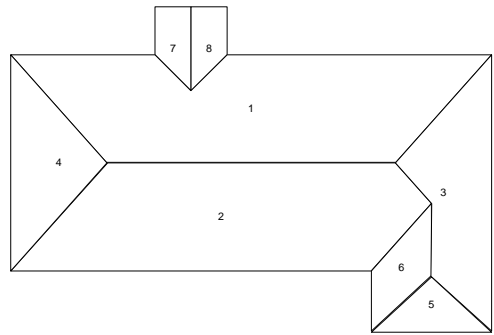

*Figure 12*: Complex Roof Layout

As in previous experiments, the greedy search clearly outperforms the other algorithms in terms of computation time and solution efficiency. Table 1 reveals a very high efficiency rate of 96.00% being achieved using this method.

Table 1

*Summary of MCPO Solution Efficiency*

|  | **Irregular Panels** | | | **Material Usage** | | | **Shared Edge Length** | | |
|---|---|---|---|---|---|---|---|---|---|
|  | GS | MC | GA | GS | MC | GA | GS | MC | GA |
| Simple Rectangle | 14 | 14 | 14 | 100.00% | 100.00% | 100.00% | 2450.00 | 2450.00 | 2450.00 |
| Single Wall | 23 | 19 | 19 | 87.81% | 84.98% | 85.42% | 2891.58 | 2397.09 | 2357.71 |
| Simple Roof | 24 | 24 | 24 | 92.29% | 77.05% | 76.40% | 1733.93 | 1613.83 | 1636.32 |
| Complex Roof | 129 | 129 | 129 | 96.00% | 65.95% | 71.23% | 9714.97 | 9387.27 | 9516.96 |

None of the optimization results produced by the Monte Carlo method achieve comparable efficiency. The same table shows its efficiency of 65.95%, nowhere near that achieved by the greedy search. The Genetic Algorithm performs somewhat better than the Monte Carlo technique this time. Solution efficiency of 71.23% has been attained, which is higher than those from MC even though still far behind the greedy search.

Similar to the preceding experiment, better efficiency is generally achieved when 180 degree rotation is an option. There is exception with GA however, where the best efficiency is found when no rotation is allowed. Once again the material constraints prove to be limiting factor for the potential of reducing the total number of panels required. Another similarity found with the previous experiment is the longer shared edge consistently obtained when best-fit placement strategy is used. The best-fit strategy also seems to contribute towards better overall efficiency, although exceptions still exist.

**Discussion**

The actual experiments were conducted with various MCPO problems with increasing complexity. More straightforward problems were used to provide empirical proof that the software actually performs the way it was designed and all the search algorithms apply their underlying logic to find this domain-specific solution. More complex problems were used to gain insight to comparative performance of the optimization algorithms, particularly in terms of computation time and solution quality. Solution quality is defined by the efficiency of material usage (or minimization of wasted material) and the length of shared edges (indicating the effort of cutting the pieces off the stock panels). Greedy search proves to be very effective by consistently outperforming other algorithms on both accounts. Greedy search has decisive advantage in computation time because the task of constructing the solution using this method is equivalent to just one fitness function evaluation on MC and GA. It also converges to a local optimum every time, of which the efficiency is always higher than that of MC and GA.

Multiple stock panels of varying size have been provided for each layout optimization problem. It has been observed that stock panels of smaller size would typically yield better efficiency, though in reality may suffer from higher installation costs. The single wall problem appears to be an anomaly where both the MC and GA implements found solutions of better efficiency with stock panel larger than that used by greedy algorithm, as indicated by the number of irregular panels. Recall that identical point of origin at (0, 0) is used for all cases, meaning the number of irregular panels resulting from the first stage solution is identical for a given stock panel dimension. The lower efficiency of the solutions however, indicates that the decision to use larger panel was sub-optimal.

The performance difference between MC and GA only slight when only a small number of pieces are involved. Both are able to find reasonable solution, compared to that achieved by greedy search, with only marginal difference in efficiency. When a larger number of irregular panels are involved, such as the case with complex roof, the GA performance advantage over MC becomes evident. These findings naturally raise a question of why such a crude algorithm can perform so much better than its much more sophisticated counterpart. Especially when compared to a GA, which is widely accepted as a powerful tool for solving multi-variable optimization class of problems to which the second-stage problem of MCPO belong.

Because the optimization algorithms have been implemented as integral parts of a computer application solving real rather than hypothetical problems, there is a number of contributing factors to be considered for the answer. The first is the relatively low level of sophistication possessed by the GA implementation. The actual implementation has been based on a simple variant of GA once coded for the benefit of students and researchers new to the subject. Such implementation is characterized by single crossover point, pair selection for breeding exclusively based on fitness value alone without regard to the actual bit patterns in the chromosome, and lack of elitism. These deficiencies alone may be directly responsible in the algorithms failure to solve complex problems.

The second fundamental problem with the use of GA in solving MCPO problem is the parameter modeling, which also applies to MC method. The concept of clustering is used in the prevailing model to address the problem of mapping the irregular panels to an undetermined number of stock panels. While the model solves this particular problem quite well, it completely disregards a host of crucial parameters to be solved in the individual nesting tasks. Relegating the actual nesting of irregular pieces from the clusters to individual stock panels to greedy search has been done as a pragmatic measure taken in the interest of generating valid nesting layout at minimum computational cost. Special provisions were also needed to effectively deal with invalid clusters. The current parameter modeling has been found far from ideal. It is quite possible that a better model may realize the true potential of the GA in solving MCPO problem. Constructing such a model however is beyond the scope of this investigation. Further interest in achieving better MCPO solutions using the GA may warrant future study in this area. On a smaller scale, various aspects of the current model, such as better realization of best-fit strategy and invalid cluster handling, can also be subject to more thorough study.

From the user perspective, the experiment results reveal that the use of novel optimization algorithms such as MC and GA has not been justified at the current stage of the software maturity. Employing the greedy search is the most logical choice for solving MCPO problems due to its low resource requirements and high quality solutions. As a commercial application, the MCPO software delivers value to the user in at least three different ways. The first is that a great amount of manual work involved in planning for panel layout project has been automated. The automated process gives the user detailed information regarding the number of stock panels required, the nesting plan for each panel, and the layout plan for the actual sections of the physical building. This wealth of information in turn allows the user to more accurately predict the costs associated with material and labor required to undertake the project.

The second benefit to the user is the optimization capability that helps him to minimize the project cost by making the necessary selection from different types of stock panel as well as making sure that minimum number of panels needs to be allocated. The amount of computational task needed to accomplish the optimization is such that manual optimization is unlikely to yield comparable result except in very simple cases.

Finally, the software capability of solving multiple container problem means that optimization does not need to be performed on the basis of individual sections of the building. As previously mentioned, the use of smaller stock panel typically results in better material efficiency. It follows that the material usage efficiency tends to improve when the ratio of container area to the stock panel area increases. Furthermore using multiple containers for a single MCPO problem is a good way of improving the ratio. The important implication is that not only solving MCPO for multiple sections of the building in a single optimization run becomes possible, but doing so actually generates less waste for the overall project than it would if the sections are optimized individually.

## Conclusions

Decomposing MCPO into a two-stage optimization model provides a solid ground for constructing a well-functioning solution. The study has also proven that with the support of appropriate analysis, software application to solve complex problem such as MCPO can be successfully implemented using standard modeling and programming tools. Successful implementation of the software in turn proves the feasibility of constructing MCPO solution automatically for commercial use with current computing technologies.

A series of experiments have demonstrated outstanding performance of greedy search in comparison with simulation-based search algorithms represented by Monte Carlo technique and Genetic Algorithm. While this result is not surprising for Monte Carlo technique given its inherent inefficiency, the unexpected lack of performance of such a sophisticated method as Genetic Algorithm calls for further investigation in the area.

There are a number of possible reasons for Genetic Algorithm's relative poor performance. The first is the efficiency of the coded implementation, which has been based of unsophisticated version of the algorithm featuring naïve strategies in accomplishing its key sub-tasks. The second possible reason is the accuracy of the parameter modeling that prevailed, in which many important parameters of the nesting problem have been omitted to be optimized by external processes. Finally, the MCPO class of problems may have certain characteristics that make Genetic Algorithm unsuitable to solve them. None of these assertions have been proved however, implying the need of further research in the area.

The use of a more sophisticated implementation of Genetic Algorithm should also be explored. Given the lack of sophistication of current implementation, it is possible that improvement in the evolution mechanism of the algorithm will have direct impact to the result of the second-stage solution. Key aspects of the algorithms that need improvement are the selection policy, crossover mechanism, and the introduction of elitism where best individuals are carried over to the succeeding generations.

Successful use of Monte Carlo technique and the Genetic Algorithm has proven that generic optimization algorithms can be used to solve the second-stage problem. The implication is that other optimization techniques could also be used in their place. Various optimization techniques such as Swarm Intelligence, Simulated Annealing, and Tabu Search can potentially increase the effectiveness of the search for the second-stage optimization.

## Acknowledgements

This research has been supported by Technology New Zealand through the Technology for Industry Fellowships scheme under grant number BISCO502

## References

Ansari, N., & Hou, E. (1997). *Computational Intelligence for Optimization*. Newark, New Jersey: Kluwer Academic Publishers.

Bäck, T., & Schwefel, H.-P. (1996, 20-22 May). Evolutionary computation: an overview. Proceedings of IEEE International Conference on Evolutionary Computation, Nagoya, Japan.

Bounsaythip, C., & Maouche, S. (1997, 12-15 Oct). Irregular shape nesting and placing with evolutionary approach. *Proceedings of the 1997 IEEE International Conference on Systems, Man, and Cybernetics ('Computational Cybernetics and Simulation')*, Orlando, FL, United States.

Bounsaythip, C., Maouche, S., & Neus, M. (1995, 22-25 Oct). Evolutionary search techniques application in automated layout-planning optimization problem. *Proceedings of the 1995 IEEE International Conference on Systems, Man and Cybernetics ('Intelligent Systems for the 21st Century',* Vancouver, BC, Canada

Connor, A.M. & Siringoringo, W.S. (2007, 7-9 Nov). Using Genetic Algorithms to solve layout optimisation problems in residential building construction, *Proceedings of the 20th International Conference on Computer Applications in Industry and Engineering*, San Francisco, California, USA.

Cormen, T., Leiserson, C., Rivest, R., & Stein, C. (2003). *Introduction to Algorithms (Second ed.)*. Cambridge, Massachusetts: The MIT Press.

Dean, T., Allen, J., & Aloimonos, Y. (1995). *Artificial Intelligence: Theory and Practice*. Menlo Park, California: Addison-Wesley Publishing.

Dyckhoff, H. (1990). Typology of cutting and packing problems. *European Journal of Operational Research*, 44(2), 145-159.

Faina, L. (1999). Application of simulated annealing to the cutting stock problem. *European Journal of Operational Research*, 114(3), 542-556.

Goldberg, D.E. (1989), Genetic algorithms in search, optimization and machine learning, Kluwer Academic Publishers, Boston, MA.

Hakimi, S. L. (1988, Jun 7-9). Problem on rectangular floorplans. *Proceedings of the 1988 IEEE International Symposium on Circuits and Systems*, Espoo, Finland.

Holland, J. *Adaptation in Natural and Artificial Systems*. Doctoral Thesis, University of Michigan Press, 1975.

Hsu, Y. C., & Kubitz, W. J. (1988). ALSO: A System for Chip Floorplan Design. Integration, *VLSI Journal*, 6(2), 127-146.

Imahori, S., Yagiura, M., & Ibaraki, T. (2005). Improved local search algorithms for the rectangle packing problem with general spatial costs. *European Journal of Operational Research*, 167(1), 48-67.

CITATION: Siringoringo, W.S., Connor, A.M., Clements, N. & Alexander, N. (2008) "Minimum cost polygon overlay with rectangular shape stock panels", International Journal of Construction Education & Research, 4(3), 1-24. DOI: 10.1080/15578770802494516

Kiyota, K., & Fujiyoshi, K. (2000, 29-31 May). Simulated annealing search through general structure floorplans using sequence-pair. *Proceedings of the IEEE 2000 International Symposium on Circuits and Systems*, Geneva, Switzerland.

Lamousin, H., & Waggenspack Jr., W. N. (1997). Nesting of two-dimensional irregular parts using a shape reasoning heuristic. *CAD Computer Aided Design*, 29(3), 221-238.

Laurent, D. G., & Iyengar, S. S. (1982). Heuristic Algorithm for Optimal Placement of Rectangular Objects. *Information Sciences*, 26(2), 127-139.

Lodi, A., Martello, S., & Monaci, M. (2002). Two-dimensional packing problems: A survey. *European Journal of Operational Research*, 141(2), 241-252.

Murata, H., Fujiyoshi, K., Nakatake, S., & Kajitani, Y. (1995, 5-9 Nov). Rectangle-packing-based module placement. *Proceedings of the 1995 IEEE/ACM International Conference on Computer-Aided Design*, San Jose, CA, USA.

Nagao, T. (1996, 20-22 May). Homogeneous Coding for Genetic Algorithm Based Parameter Optimization. *Proceedings of the IEEE Conference on Evolutionary Computation*, Nagoya, Japan.

Oloffson, P. (2005). *Probability, Statistics, and Stochastic Processes*. Houston, Texas: Wiley-Interscience.

Sibley-Punnett, L., & Bossomaier, T. (2001, 19-21 Aug). Optimisation techniques for roof layout. *Proceedings of IEEE Region 10 International Conference on Electrical and Electronic Technology ('TENCOM')*, Singapore.

Siringoringo, W.S. (2007) Minimum cost polygon overlay with rectangular shape stock panels, Masters Thesis, Auckland University of Technology, Auckland, New Zealand.

Weisstein, E. W. (1999). Monte Carlo Method. *Retrieved June, 2006, from http://mathworld.wolfram.com/MonteCarloMethod.html*